\documentclass[11pt]{article}

\usepackage[preprint]{acl}

\newcommand{\acceptednote}[1]{%
  \begingroup
  \renewcommand\thefootnote{}%
  \footnotetext{#1}%
  \endgroup
}

\usepackage{times}
\usepackage{latexsym}

\usepackage[T1]{fontenc}

\usepackage[utf8]{inputenc}

\usepackage{microtype}

\usepackage{inconsolata}
\usepackage{amsmath}
\usepackage{graphicx}

\usepackage{xspace}

\usepackage{amsthm}
\usepackage{amssymb}

\usepackage{algorithm}
\usepackage{algorithmic}
\usepackage{array}
\usepackage{enumitem}
\setlist[itemize]{leftmargin=*,itemsep=0cm,topsep=0.2cm}
\usepackage[table]{xcolor}

\usepackage{booktabs}
\usepackage{multirow}
\usepackage{makecell}
\usepackage{threeparttable}
\usepackage{graphicx}
\usepackage{bbm}
\usepackage{pifont}
\usepackage{stfloats}
\usepackage{enumitem}
\newcommand{\cmark}{\ding{51}}
\newcommand{\xmark}{\ding{55}}

\usepackage{pifont}
\usepackage{xcolor}
\usepackage{tabularx}
\usepackage{hyperref}
\newcommand{\pmark}{\textcolor{orange!80!black}{\(\triangle\)}}

\title{TRACE: Single-Pass Decoding-Trace Risk Localization for Generation Calibration}

\author{
Yuebin Xu, ~Xuemei Peng, ~Junlan Chen,
~Zhiyi Chen, ~Zeyi Wen\thanks{Corresponding author} \\
The Hong Kong University of Science and Technology (Guangzhou)\\
\texttt{\{yxu349,xpeng558,jchen421,zchen986,wenzeyi\}@hkust-gz.edu.cn}
\hypersetup{
  pdftitle={TRACE: Single-Pass Decoding-Trace Risk Localization for Generation Calibration},
  pdfauthor={Yuebin Xu, Xuemei Peng, Junlan Chen, Zhiyi Chen, and Zeyi Wen}
}
}
\begin{document}
\maketitle
\acceptednote{\textit{Accepted to Findings of the Association for Computational Linguistics: EMNLP 2026.}}

\begin{abstract}
Reliable confidence estimation is essential for large language model deployment. However, answer-level calibration remains challenging because generation errors are often localized: a response may be fluent and high-probability overall while still failing at a critical number, entity, or factual claim. Existing estimators compress token probabilities, sequence likelihoods, entropy, or beam statistics into a global score, which can dilute such local risk signals. We propose \textsc{TRACE}, a single-pass, decoded-answer-preserving confidence estimator that treats decoding-time uncertainty as a trajectory through three steps: (i) recording token-level surprisal and predictive entropy during decoding, (ii) applying local risk operators to preserve uncertainty spikes, and (iii) converting localized trace risk into answer-level confidence. TRACE produces a label-free risk score, while TRACE+ calibrates trace-only features into probabilities using a held-out split, without extra generations or external verifiers. We evaluate four tasks against 19 calibration baselines, and TRACE+ reduces Brier from $0.149$ to $0.137$ and improves AUROC from $0.758$ to $0.792$ over the strongest likelihood baseline. Across seven LLMs, TRACE+ improves over the best non-TRACE baseline pool from $0.136$ to $0.120$ Brier and from $0.764$ to $0.817$ AUROC. Results show localizing decoding-time risk provides a general approach to calibration.
\end{abstract}

\section{Introduction}

Large language models (LLMs) are increasingly deployed in generation settings where users need not only an answer, but also a reliable estimate of whether that answer is correct~\citep{guo2017calibration,desai2020calibration,jiang2021can}.
This problem is more challenging than standard classification calibration: a generated answer is a sequence of tokens, correctness may depend on a single number, entity, or factual claim, and a fluent response can still be wrong
\citep{kadavath2022language,manakul2023selfcheckgpt,kuhn2023semantic}. We study \emph{general generation calibration}: estimating the answer-level correctness probability of a decoded response across diverse generation tasks.

A common approach is to compress the decoded answer into a single global confidence score, such as maximum token probability, mean token probability, length-normalized log-probability, total sequence likelihood, or average token
entropy~\citep{guo2017calibration,yona2022useful,aichberger2024rethinking}. Recent single-pass and near-single-pass methods further exploit beam-distribution statistics~\citep{flores2025improving} or semantic reweighting
\citep{duan2024shifting,bakman2024mars}. While effective in many cases, these methods largely summarize a generation as a sequence-level statistic. This can erase an important part of the decoding process: \emph{where} uncertainty occurs.
A response may appear confident on average while becoming locally unstable at the critical step where a wrong digit, entity, or factual attribution is produced.

Multi-sample uncertainty estimators address generation uncertainty from another direction by comparing multiple sampled outputs or semantic equivalence classes \citep{manakul2023selfcheckgpt,kuhn2023semantic}. However, these methods require additional generations and may change the inference workflow, which can be costly when confidence estimates are needed for every answer. We therefore focus on a stricter and more deployment-oriented setting: \emph{single-pass, decoded-answer-preserving confidence estimation}. In this setting, all methods score the same decoded answer and do not use extra samples, retrieval, external verifiers, or LLM-as-a-judge labels at actual inference time.

\begin{figure*}[ht]
    \centering
    \includegraphics[width=\textwidth]{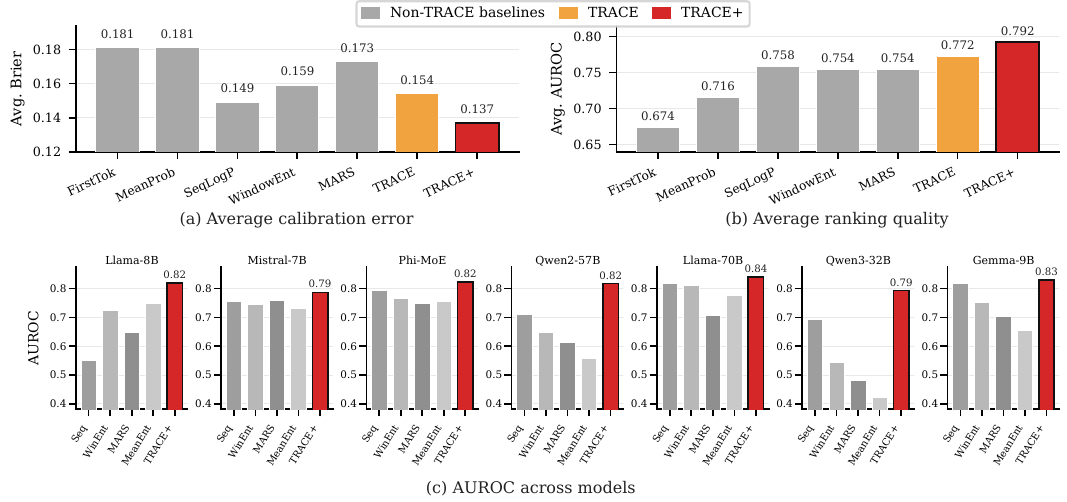}
\caption{
Main experimental results.
(a) Average Brier score across MLQA, SVAMP, TriviaQA, and TruthfulQA, where lower is better.
(b) Average AUROC across the same tasks, where higher is better.
(c) Cross-model AUROC in the main experiments.
Gray, orange, and red bars denote representative non-TRACE baselines, TRACE, and TRACE+, respectively.
TRACE improves over standard sequence-level baselines, and TRACE+ achieves the strongest overall performance across both tasks and model families.
}
    \label{fig:trace_performance_teaser}
\end{figure*}

Our key observation is that decoding-time uncertainty should be treated as a token-level trajectory rather than immediately collapsed into a global score. Local entropy, surprisal, and risk peaks can appear near the critical answer span in real errors, such as a wrong number in arithmetic reasoning, a wrong entity in open-domain QA, or an unsupported factual claim in truthfulness-sensitive generation. These local peaks are not direct labels of token-level error, but they are informative indicators of answer-level risk. Figure~\ref{fig:trace_performance_teaser} provides an overview of the empirical gains from this trace-based design across tasks and model families.

Based on this observation, we propose \textsc{TRACE}, a single-pass confidence estimator based on \emph{decoding-trace risk localization}. Given a fixed decoded answer, TRACE records token-level surprisal and predictive entropy during the original generation pass, but instead of immediately averaging them, it applies local risk operators that preserve position-sensitive uncertainty spikes. We further introduce TRACE+, which maps trace-only features to calibrated probabilities using a lightweight calibrator trained on a held-out calibration split. Thus, TRACE is label-free, while TRACE+ uses supervised post-hoc calibration requiring only decoding-trace features at inference.

We evaluate TRACE and TRACE+ on multilingual QA, arithmetic reasoning, open-domain QA, and truthfulness-sensitive generation. Across four tasks, 19 confidence estimators, and seven LLMs, TRACE+ achieves the best average Brier score and AUROC, transfers across model families and scales, and remains strong without conventional selected-token likelihood summaries. Our contributions are:

\begin{itemize}
    \item We formalize and evaluate generation calibration under a single-pass protocol, where all methods score the same generated answer without extra samples, retrieval, or external verifiers.

    \item We propose \textsc{TRACE} and TRACE+, which estimate answer-level confidence from localized surprisal and entropy patterns in the decoding trace, rather than relying on global sequential summaries that dilute critical uncertainty spikes.
    
    \item We evaluate across four tasks, 19 confidence-estimating baselines, and seven LLMs, showing that TRACE+ achieves the best average Brier score and AUROC across diverse models.
    
\end{itemize}

\begin{figure*}[t]
    \centering
    \includegraphics[width=\textwidth]{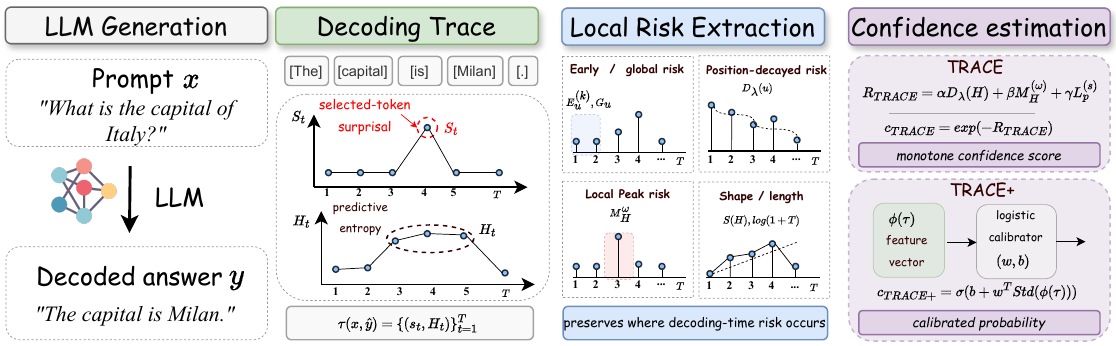}
    \caption{
    Overview of TRACE and TRACE+.
    Given a prompt and a decoded answer, TRACE extracts token-level surprisal and predictive entropy from the decoding trace, localizes risk through early/global, position-decayed, local-peak, and shape/length operators, and converts the resulting risk into a monotone confidence score.
    TRACE+ further standardizes trace features and applies a lightweight logistic calibrator to produce a calibrated probability.
    }
    \label{fig:trace_framework}
\end{figure*}

\section{Related Work}
\paragraph{Calibration and single-pass confidence.}
Confidence estimation is commonly framed as calibration, where predicted confidence should match empirical correctness~\citep{platt1999probabilistic,niculescu2005predicting,naeini2015obtaining,guo2017calibration,kuleshov2018accurate}. In NLP, language models and QA systems are often miscalibrated, especially under distribution shift or when probabilities are used directly as confidence~\citep{desai2020calibration,jiang2021can,kadavath2022language}. Generation confidence estimators therefore derive answer-level scores from decoding statistics, including maximum softmax probability~\citep{hendrycks2017baseline,guo2017calibration}, mean/minimum token probability, length-normalized log-probability, sequence likelihood, and entropy~\citep{yona2022useful,malinin2021uncertainty,aichberger2024rethinking}. Recent methods further use first-token confidence~\citep{chen2026firsttoken}, entropy dynamics~\citep{zhu2026edis}, uncertainty patterns over generated sequences~\citep{manakul2023selfcheckgpt}, and beam-distribution statistics~\citep{flores2025improving}. These estimators are efficient, but many compress decoding into a global sequence-level score; TRACE instead preserves localized surprisal and entropy patterns along the decoding trajectory.

\paragraph{Multi-sample and semantic uncertainty.}
Another line of work estimates uncertainty through self-evaluation or semantic comparison. Self-consistency and sampling-based methods use agreement across completions as reliability evidence~\citep{wang2023selfconsistency,manakul2023selfcheckgpt}, while semantic entropy measures uncertainty over meaning-level equivalence classes~\citep{kuhn2023semantic}. Related work also studies verbalized confidence, self-evaluation, and whether models can assess the truth of their own answers~\citep{kadavath2022language,lin2022teaching,tian2023just,xiong2024can}. These methods can reveal uncertainty beyond a single decoded answer, but require extra samples or inference procedures. TRACE instead targets a decoded-answer-preserving setting without extra sampling, retrieval, external verifiers, or LLM-as-a-judge labels. It is also related to TokenSAR and MARS, which reweight tokens or sentences by semantic relevance or meaning contribution~\citep{duan2024shifting,bakman2024mars}; however, TRACE asks where decoding-time risk becomes locally concentrated rather than which generated components are semantically important.

\section{Method}
\subsection{Problem Formulation}
Given a decoded answer $\hat{y}$ for prompt $x$, we estimate an answer-level confidence score $c(x,\hat{y})$, where larger values indicate higher likelihood of correctness, following standard confidence-calibration practice
\citep{guo2017calibration,desai2020calibration,jiang2021can}. We focus on a decoded-answer-preserving setting: all methods score the same $\hat{y}$ after generation, without extra samples, retrieval, or external verifiers. Let $z\in\{0,1\}$ denote task-specific correctness. TRACE is label-free, while TRACE+ uses labels only on a held-out calibration split to map trace-derived features to calibrated probabilities,
\[
    \hat{p}(z=1\mid x,\hat{y}) = g(r(x,\hat{y})) .
\]
This follows standard post-hoc calibration practice \citep{platt1999probabilistic,zadrozny2002transforming,niculescu2005predicting}. For Brier and ECE, TRACE and all baselines use the same held-out calibration protocol \citep{brier1950verification,naeini2015obtaining}.

\begin{figure}[t]
    \centering
    \includegraphics[width=\columnwidth]{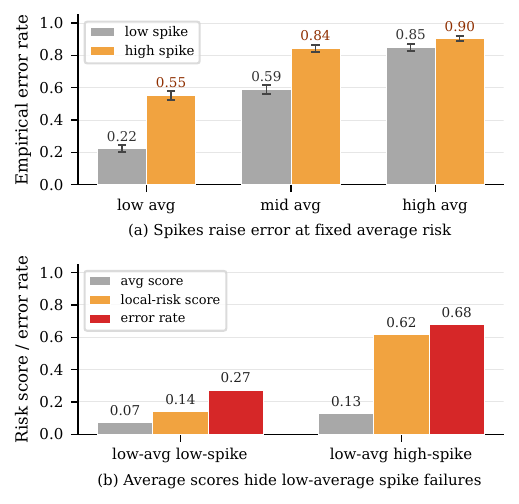}
    \caption{
    Token-level spike evidence for TRACE.
    (a) At comparable average-risk levels, high-spike samples have substantially
    higher empirical error rates than low-spike samples.
    (b) Among low-average-risk samples, average-based scores remain low even when
    local-risk scores and empirical error rates rise sharply.
    }
    \label{fig:trace_spike_evidence}
\end{figure}

\subsection{Method Overview}

Figure~\ref{fig:trace_framework} summarizes the TRACE pipeline.
Given a prompt and a fixed decoded answer, TRACE proceeds in four steps.
\textbf{(i) LLM generation.}
The model generates an answer under the original decoding policy; TRACE keeps this answer fixed.
\textbf{(ii) Decoding trace construction.}
During the same pass, TRACE records selected-token surprisal and predictive entropy at each decoding step.
\textbf{(iii) Local risk extraction.}
TRACE applies early/global, position-decayed, local-peak, and shape/length operators to preserve localized uncertainty.
\textbf{(iv) Confidence estimation.}
TRACE produces a label-free monotone confidence score, while TRACE+ calibrates trace-derived features into answer-level probabilities using a held-out split.

\subsection{Decoding Traces as Local Risk Signals}

Global confidence estimators such as mean token probability, mean entropy, or sequence likelihood collapse a generated answer into a single scalar. While this captures overall uncertainty, it discards \emph{where} uncertainty occurs during
decoding. This location information is often decisive: an answer can appear fluent overall yet fail at a final digit, an entity name, or one unsupported factual claim.

Figure~\ref{fig:trace_spike_evidence} motivates preserving local decoding structure: within bins of similar average risk, stronger local spikes correspond to higher empirical error rates, and low-average but high-spike examples are
underestimated by average-based scores. These diagnostic results show that local instability provides risk information beyond global summaries alone (Appendix~\ref{app:trace_spike_evidence}). We therefore model decoding-time
uncertainty as an ordered trajectory rather than collapsing it into a sequence-level statistic.

From the original generation pass, without extra generation and without changing the decoded answer, TRACE records two token-level signals at each decoding step $t$ under $P_\theta(\cdot\mid x,\hat{y}_{<t})$. The first is selected-token
surprisal,

\[
    s_t=-\log P_\theta(\hat{y}_t\mid x,\hat{y}_{<t}),
\]
which measures uncertainty about the token actually generated. The second is predictive entropy,
\[
    H_t=-\sum_{v\in\mathcal{V}}
    P_\theta(v\mid x,\hat{y}_{<t})
    \log P_\theta(v\mid x,\hat{y}_{<t}),
\]
which measures how diffuse the full candidate distribution is. The token-level decoding trace is
\[
    \tau(x,\hat{y})=\{(s_t,H_t)\}_{t=1}^{T}.
\]
TRACE uses this ordered trace to preserve where decoding-time risk occurs.

\subsection{TRACE: From Local Risk to Confidence}

TRACE converts the token-level decoding trace into an answer-level confidence score while preserving localized risk. Rather than averaging the entire trace into one global statistic, TRACE combines three complementary risk operators that capture different failure modes: position-decayed entropy risk, maximum local-window entropy risk, and
length-normalized selected-token surprisal:
\begin{equation}
\begin{aligned}
R_{\text{\textsc{TRACE}}}(x,\hat{y})
&= \alpha D_\lambda(H)
 + \beta M_H^{(w)}
 + \gamma L_\rho(s), \\
L_\rho(s)
&= \frac{\sum_{t=1}^{T}s_t}{T^\rho}.
\end{aligned}
\end{equation}

Here, $D_\lambda(H)$ emphasizes uncertainty at position-dependent decoding steps, $M_H^{(w)}$ preserves the highest local-window entropy risk, and $L_\rho(s)$ captures selected-token surprisal with length normalization. The coefficients $\alpha,\beta,\gamma\ge0$ satisfy $\alpha+\beta+\gamma=1$; $\lambda$, $w$, and $\rho$ control the decay rate, local window size, and length normalization. We use one fixed hyperparameter configuration across tasks and models, without fitting these hyperparameters on evaluation labels. TRACE converts risk to a monotone confidence score:
\[
    c_{\textsc{TRACE}}(x,\hat{y})
    =
    \exp(-R_{\textsc{TRACE}}(x,\hat{y})).
\]
This score is used directly for ranking metrics and should not be interpreted as a calibrated probability before post-hoc calibration. Full feature definitions are provided in Appendix~\ref{app:trace_features}.

\subsection{TRACE+: Calibrated Trace Confidence}

TRACE+ learns how to map trace-derived risk patterns to calibrated probabilities. For each decoded answer, we construct a trace-only feature vector $\boldsymbol{\phi}(\tau)$ from the same decoding trace, including early and global entropy confidence, answer length, entropy slope, and position-decayed entropy and surprisal confidence. We intentionally exclude conventional selected-token likelihood summaries, such as minimum token probability, geometric mean token probability, and low-probability token rate; these are included only in the ``+ seq. likelihood'' ablation. We train a lightweight logistic calibrator on a held-out calibration split:
\[
    c_{\textsc{TRACE+}}(x,\hat{y})
    =
    \sigma\!\left(
    b+\mathbf{w}^{\top}\mathrm{Std}(\boldsymbol{\phi}(\tau))
    \right),
\]
where $\mathrm{Std}(\cdot)$ standardizes each feature using calibration-split statistics. The calibrator is not an answer verifier, does not inspect the answer semantically, and does not use alternative generations. TRACE and TRACE+ require no additional decoding passes and use only token-level probabilities and entropies from the original generation.

\section{Experiments}

\label{sec:experiments}

\begin{table*}[t]
\centering
\scriptsize
\setlength{\tabcolsep}{2.2pt}
\renewcommand{\arraystretch}{1.04}
\resizebox{\textwidth}{!}{%
\begin{tabular}{@{}lcccccccccc@{}}
\toprule
\multirow{2}{*}{Method} &
\multicolumn{2}{c}{MLQA} &
\multicolumn{2}{c}{SVAMP} &
\multicolumn{2}{c}{TriviaQA} &
\multicolumn{2}{c}{TruthfulQA} &
\multicolumn{2}{c}{Avg.} \\
\cmidrule(lr){2-3}
\cmidrule(lr){4-5}
\cmidrule(lr){6-7}
\cmidrule(lr){8-9}
\cmidrule(lr){10-11}
& Brier $\downarrow$ & AUROC $\uparrow$
& Brier $\downarrow$ & AUROC $\uparrow$
& Brier $\downarrow$ & AUROC $\uparrow$
& Brier $\downarrow$ & AUROC $\uparrow$
& Brier $\downarrow$ & AUROC $\uparrow$ \\
\midrule

\multicolumn{11}{@{}c}{\textit{Token- and sequence-level confidence baselines}} \\
First-Token Prob~\citep{chen2026firsttoken}
& 0.230 & 0.580 & 0.211 & 0.721 & 0.188 & 0.806 & 0.094 & 0.590 & 0.181 & 0.674 \\
MeanProb~\citep{flores2025improving}
& 0.228 & 0.692 & 0.196 & 0.782 & 0.206 & 0.774 & 0.094 & 0.615 & 0.181 & 0.716 \\
Len-Norm LogP~\citep{bakman2024mars}
& 0.226 & 0.694 & 0.192 & 0.778 & 0.198 & 0.783 & 0.094 & 0.619 & 0.178 & 0.719 \\
SeqLogP / Total NLL~\citep{aichberger2024rethinking}
& \underline{0.174} & \textbf{0.812} & 0.146 & 0.791 & 0.182 & 0.811 & 0.093 & 0.619 & \underline{0.149} & 0.758 \\
MinProb~\citep{manakul2023selfcheckgpt}
& 0.214 & 0.741 & 0.184 & 0.745 & 0.186 & 0.795 & 0.094 & 0.585 & 0.169 & 0.717 \\
Mean Token Entropy~\citep{flores2025improving}
& 0.222 & 0.697 & 0.170 & 0.816 & 0.188 & 0.796 & 0.093 & 0.626 & 0.168 & 0.734 \\
Max Token Entropy~\citep{manakul2023selfcheckgpt}
& 0.218 & 0.745 & 0.182 & 0.763 & 0.180 & 0.820 & 0.093 & 0.613 & 0.168 & 0.735 \\
High-Prob Token Rate~\citep{szymczyk2026monitoring}
& 0.227 & 0.663 & 0.178 & 0.751 & 0.214 & 0.719 & 0.093 & 0.610 & 0.178 & 0.686 \\
First-$k$ Entropy~\citep{chen2026firsttoken}
& 0.218 & 0.688 & 0.170 & 0.811 & 0.182 & 0.806 & 0.093 & 0.629 & 0.166 & 0.734 \\
Last-$k$ Entropy~\citep{li2026semantic}
& 0.227 & 0.600 & 0.170 & 0.801 & 0.212 & 0.722 & 0.094 & 0.553 & 0.176 & 0.669 \\

\midrule
\multicolumn{11}{@{}c}{\textit{Trajectory, beam-distribution, and semantic baselines}} \\
Windowed Entropy~\citep{sriramanan2024llm}
& 0.205 & 0.753 & 0.164 & 0.817 & 0.176 & 0.822 & 0.093 & 0.625 & 0.159 & 0.754 \\
Entropy Variance~\citep{zhu2026edis}
& 0.221 & 0.706 & 0.201 & 0.713 & 0.198 & 0.775 & 0.093 & 0.624 & 0.178 & 0.705 \\
Entropy Slope~\citep{zhu2026edis}
& 0.234 & 0.445 & 0.202 & 0.351 & 0.196 & 0.235 & 0.094 & 0.435 & 0.181 & 0.366 \\
Token-Prob Slope~\citep{shapiro2026halt}
& 0.232 & 0.434 & 0.206 & 0.360 & 0.205 & 0.254 & 0.094 & 0.439 & 0.184 & 0.372 \\
Beam-Ratio~\citep{flores2025improving}
& 0.223 & 0.634 & 0.137 & 0.850 & 0.219 & 0.701 & 0.094 & 0.537 & 0.168 & 0.681 \\
Beam-TailThinness~\citep{flores2025improving}
& 0.219 & 0.682 & 0.132 & \textbf{0.887} & 0.208 & 0.738 & 0.094 & 0.550 & 0.163 & 0.714 \\
Beam Entropy~\citep{flores2025improving}
& 0.220 & 0.674 & \underline{0.131} & \underline{0.885} & 0.209 & 0.734 & 0.094 & 0.550 & 0.164 & 0.711 \\
TokenSAR~\citep{duan2024shifting}
& 0.225 & 0.659 & 0.185 & 0.841 & 0.193 & 0.793 & 0.093 & \textbf{0.661} & 0.174 & 0.739 \\
MARS~\citep{bakman2024mars}
& 0.224 & 0.726 & 0.184 & 0.843 & 0.189 & 0.802 & 0.093 & 0.645 & 0.173 & 0.754 \\

\midrule
TRACE
& 0.197 & 0.786 & 0.161 & 0.825 & \underline{0.168} & \underline{0.835} & \underline{0.092} & 0.641 & 0.154 & \underline{0.772} \\
TRACE+
& \textbf{0.173} & \underline{0.803} & \textbf{0.124} & 0.864 & \textbf{0.160} & \textbf{0.847} & \textbf{0.091} & \underline{0.654} & \textbf{0.137} & \textbf{0.792} \\
\bottomrule
\end{tabular}%
}
\caption{Main comparison on general generation calibration. All methods score the same decoded answer; beam-based methods additionally use beam statistics from the same prompt.}
\label{tab:main_generation_calibration}

\end{table*}

\subsection{Experimental Setup}
\label{sec:experiment_setup}

\paragraph{Tasks and Models.}
We evaluate general generation calibration on MLQA~\citep{lewis2020mlqa}, SVAMP~\citep{patel2021nlp}, TriviaQA~\citep{joshi2017triviaqa}, and TruthfulQA~\citep{lin2022truthfulqa}, covering multilingual QA, arithmetic reasoning, open-domain QA, and truthfulness-sensitive generation. We use Qwen2.5-7B-Instruct~\citep{yang2024qwen25} for the main comparison and seven additional LLMs for generalizability and cross-model evaluation: Llama-3.1-8B/70B-Instruct~\citep{dubey2024llama}, Mistral-7B-Instruct-v0.3~\citep{jiang2023mistral}, Phi-3.5-MoE-Instruct~\citep{abdin2024phi}, Qwen2-57B-A14B~\citep{yang2024qwen2}, Qwen3-32B~\citep{qwen2025qwen3}, and Gemma-2-9B~\citep{team2024gemma}. We report Brier score~\citep{brier1950verification}, ECE~\citep{naeini2015obtaining}, and AUROC~\citep{fawcett2006introduction}; Brier and ECE evaluate probability calibration, while AUROC measures confidence ranking and selective-prediction risk on shared decoded outputs across generation settings.

\paragraph{Baselines and Implementation Details.}
We compare TRACE with 19 baselines from five families: (i) token/sequence likelihood scores, including First-Token Prob~\citep{chen2026firsttoken}, MeanProb, length-normalized LogP, SeqLogP / Total NLL, MinProb, and high-probability token rate~\citep{szymczyk2026monitoring,aichberger2024rethinking,bakman2024mars}; (ii) entropy/position scores, including mean/max entropy, first-$k$ entropy~\citep{chen2026firsttoken}, and last-$k$ entropy~\citep{li2026semantic}; (iii) trajectory diagnostics, including windowed entropy~\citep{sriramanan2024llm}, entropy variance/slope~\citep{zhu2026edis}, and token-probability slope~\citep{shapiro2026halt}; (iv) near-single-pass beam scores, including Beam-Ratio, Beam-TailThinness, and Beam Entropy~\citep{flores2025improving}; and (v) semantic reweighting scores, including TokenSAR~\citep{duan2024shifting} and MARS~\citep{bakman2024mars}. All methods score the same decoded answer; beam methods only add beam statistics from the same prompt. AUROC uses raw ranking scores, Brier uses held-out calibrated probabilities, and results are averaged over 20 random calibration/evaluation splits. Table~\ref{tab:protocol_comparison} summarizes the main inference assumptions, and Appendix~\ref{app:baseline_protocol_details} gives the full interpretation of comparison and implementations.

\begin{figure*}[t]
    \centering
    \includegraphics[width=\textwidth]{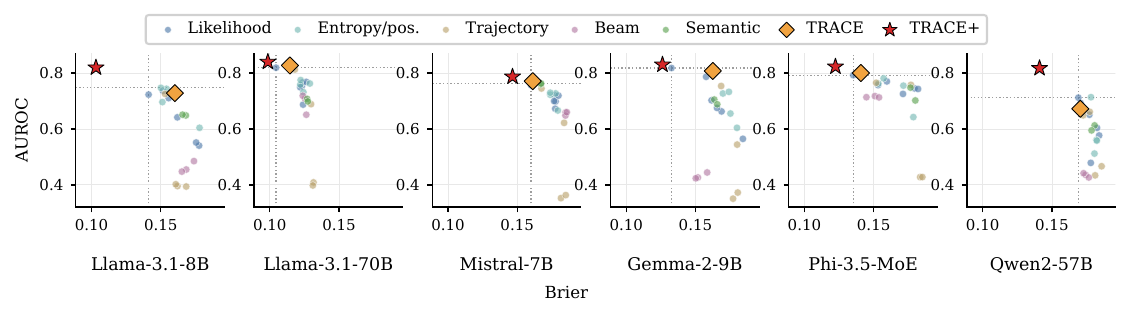}
\caption{
Cross-model estimator landscape.
Each point denotes one estimator averaged over SVAMP, TriviaQA, and TruthfulQA.
Upper-left is better; dotted lines mark the best non-TRACE Brier and AUROC.
}
    \label{fig:model_method_scatter_multiples}
\end{figure*}

\subsection{Experimental Results}
\label{sec:main_results}

\paragraph{Main results.}
Table~\ref{tab:main_generation_calibration} compares TRACE+ with likelihood-, entropy-, trajectory-, beam-, and semantic-reweighting confidence estimators on four generation tasks. We make two observations.
\textbf{(i)} TRACE+ achieves the best average calibration and discrimination performance, obtaining the lowest average Brier score and the highest average AUROC among all methods. Compared with the strongest non-TRACE method on the average columns, SeqLogP / Total NLL, TRACE+ reduces average Brier from $0.149$ to $0.137$ and improves average AUROC from $0.758$ to $0.792$. TRACE+ also attains the best Brier score on all four tasks and the best AUROC on TriviaQA; on MLQA and TruthfulQA, its AUROC is second-best, while on SVAMP it trails beam-distribution methods in AUROC but still gives the lowest Brier score. \textbf{(ii)} Strong baselines are task-dependent: SeqLogP / Total NLL performs best on MLQA, beam-distribution statistics are strongest on SVAMP, and semantic reweighting helps on TruthfulQA, but none is consistently best across tasks. TruthfulQA is highly imbalanced ($86/817$ correct), for which a constant base-rate predictor obtains a Brier score of $0.0942$; hence Brier differences are compressed despite larger AUROC differences.
TRACE+ is more stable because it aggregates localized decoding-trace evidence under the same decoded-answer setting, yielding consistently stronger average performance overall without changing the model outputs.

\begin{table*}[t!]
\centering
\scriptsize
\setlength{\tabcolsep}{3.7pt}
\renewcommand{\arraystretch}{1.06}
\resizebox{\textwidth}{!}{%
\begin{tabular}{@{}lc cc cc cc cc@{}}
\toprule
\multirow{2}{*}{Model} &
\multirow{2}{*}{Avg. Acc.} &
\multicolumn{2}{c}{Best Non-TRACE} &
\multicolumn{2}{c}{TRACE+} &
\multicolumn{2}{c}{Improvement} &
\multicolumn{2}{c}{Task Wins} \\
\cmidrule(lr){3-4}
\cmidrule(lr){5-6}
\cmidrule(lr){7-8}
\cmidrule(l){9-10}
&
& Brier $\downarrow$ & AUROC $\uparrow$
& Brier $\downarrow$ & AUROC $\uparrow$
& $\Delta$Brier $\uparrow$ & $\Delta$AUROC $\uparrow$
& Brier & AUROC \\
\midrule

Llama-3.1-8B-Instruct
& 0.396
& 0.141 & 0.747
& \textbf{0.102} & \textbf{0.819}
& +0.039 & +0.072
& 3/3 & 3/3 \\

Mistral-7B-Instruct-v0.3
& 0.438
& 0.160 & 0.763
& \textbf{0.146} & \textbf{0.790}
& +0.014 & +0.027
& 3/3 & 2/3 \\

Phi-3.5-MoE-Instruct
& 0.431
& 0.135 & 0.793
& \textbf{0.121} & \textbf{0.827}
& +0.014 & +0.034
& 3/3 & 3/3 \\

Qwen2-57B-A14B
& 0.371
& 0.169 & 0.714
& \textbf{0.143} & \textbf{0.819}
& +0.026 & +0.105
& 3/3 & 2/3 \\

Llama-3.1-70B-Instruct
& 0.592
& 0.104 & 0.819
& \textbf{0.098} & \textbf{0.844}
& +0.007 & +0.024
& 3/3 & 2/3 \\

Qwen3-32B
& 0.242
& 0.113 & 0.693
& \textbf{0.108} & \textbf{0.791}
& +0.005 & +0.097
& 3/3 & 3/3 \\

Gemma-2-9B
& 0.454
& 0.133 & 0.818
& \textbf{0.126} & \textbf{0.832}
& +0.007 & +0.014
& 3/3 & 2/3 \\

\midrule
Overall
& 0.418
& 0.136 & 0.764
& \textbf{0.120} & \textbf{0.817}
& +0.016 & +0.053
& 21/21 & 17/21 \\

\bottomrule
\end{tabular}%
}
\caption{
Cross-model generalization against the full non-TRACE baseline pool.
Metrics are averaged over SVAMP, TriviaQA, and TQA-Gen.
Best Non-TRACE reports the strongest value achieved by any non-TRACE estimator.
}
\label{tab:model_generalization}
\end{table*}

\paragraph{Broader UQ comparison.}
We further compare TRACE with broader uncertainty-estimation paradigms, including LARS~\citep{yaldiz2025}, P(True)~\citep{kadavath2022language}, verbalized confidence~\citep{tian2023just}, and an internal-state probe~\citep{ji2024}. TRACE achieves the strongest average AUROC among methods requiring neither extra inference, hidden-state access, nor labeled training, while TRACE+ achieves the best overall Brier score and AUROC. Full results and inference requirements are in Appendix~\ref{app:broader_uq}.

\paragraph{Cross-model generalization.}
Table~\ref{tab:model_generalization} evaluates TRACE+ across seven LLMs against the full competitive non-TRACE baseline pool. Best Non-TRACE is a per-metric oracle separately over all available non-TRACE estimators, averaged over SVAMP, TriviaQA, and TQA-Gen, so TRACE+ is compared with the strongest available baseline rather than a fixed method. Overall, TRACE+ reduces average Brier from $0.136$ to $0.120$ and improves AUROC from $0.764$ to $0.817$, with task-level gains in $21/21$ Brier and $17/21$ AUROC settings. Figure~\ref{fig:model_method_scatter_multiples} shows the estimator landscape varying model families, where each point is one estimator averaged over the three tasks. TRACE+ lies near the upper-left frontier, while different baseline families occupy model-dependent regions. Together, the table and figure show that TRACE+ reliably transfers across model families and scales. 

\subsection{Analysis and Ablations}
\begin{figure}[t]
    \centering
    \includegraphics[width=0.48\textwidth]{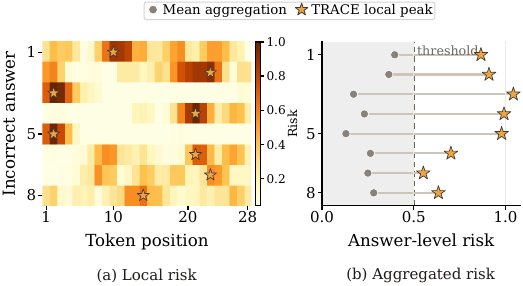}
    \caption{
    Local-risk aggregation analysis.
    Panel (a) shows real incorrect generations with localized decoding-risk peaks.
    Panel (b) compares mean aggregation with TRACE local-peak risk on the same examples.
    }
    \label{fig:trace_landscape_aggregation}
\end{figure}

\paragraph{Local-risk aggregation analysis.}
Figure~\ref{fig:trace_landscape_aggregation} examines whether TRACE captures risk that is lost by global aggregation.
We select real incorrect generations from the main evaluation set whose decoding traces contain localized high-risk regions, and compute two scores on the same traces: mean aggregation, which averages token-level risk over the whole answer, and TRACE local-peak risk, which preserves the largest localized risk region. As shown in Figure~\ref{fig:trace_landscape_aggregation}(a), many incorrect answers contain sparse risk peaks rather than uniformly high uncertainty. Figure~\ref{fig:trace_landscape_aggregation}(b) shows that mean aggregation often assigns these examples modest answer-level risk because the peak is diluted by many low-risk tokens, whereas TRACE local-peak risk remains high.
This supports our central hypothesis that answer-level failures can be locally concentrated in the decoding trajectory, and that preserving local risk structure provides information beyond global likelihood or entropy summaries.

\paragraph{Localized-error analysis.}
We further evaluate whether TRACE is particularly effective on localized errors, where correctness depends on a compact semantic span. We distinguish these from globally diffuse failures and match each error slice with correct answers by task and answer length. TRACE+ improves localized-error AUROC from $0.778$ for the strongest non-TRACE baseline to $0.805$, while TRACE reaches $0.796$. Across number/arithmetic, entity/span, and factual-claim errors, TRACE+ obtains AUROC of $0.927$, $0.836$, and $0.607$, respectively. These results show that localized decoding-trace uncertainty is informative when correctness hinges on a compact critical span. Detailed results are provided in Appendix~\ref{app:localized_error_analysis}.

\begin{table}[t]
\centering
\small
\begin{threeparttable}
\begin{tabular*}{\linewidth}{@{\extracolsep{\fill}}lccc}
\toprule
Variant & Brier $\downarrow$ & AUROC $\uparrow$ & ECE $\downarrow$ \\
\midrule
TRACE (raw)             & 0.182 & 0.772 & 0.163 \\
TRACE + scalar calib.   & 0.154 & 0.772 & 0.092 \\
Learned $(D,M,L)$       & 0.140 & 0.778 & \textbf{0.054} \\
TRACE+                  & \textbf{0.137} & \textbf{0.792} & \textbf{0.054} \\
\bottomrule
\end{tabular*}
\caption{
Bridge from TRACE to TRACE+, comparing calibration, learned operators, and richer trace features.
}
\label{tab:trace_bridge}
\end{threeparttable}
\end{table}

\paragraph{TRACE operator analysis.}
We next examine the contribution of the three risk operators in the TRACE score. Using position-decayed entropy $D$, local-window entropy $M$, or length-normalized surprisal $L$ alone gives average AUROC of $0.751$, $0.754$, and $0.758$, respectively, while their fixed combination improves to $0.772$. A learned combination of the three operators reaches $0.776$, indicating that the fixed TRACE formulation captures most of the benefit without task-specific fitting. Controlled perturbations show that disrupting risk positions reduces ranking quality, supporting position-sensitive local aggregation. Performance remains stable across broad variations of $\alpha$, $\beta$, $\gamma$, $\lambda$, $w$, and $\rho$. Full results are provided in Appendix~\ref{app:trace_operator_analysis}.

\paragraph{From TRACE to TRACE+.}
We further examine whether TRACE+ gains arise only from calibrating the fixed TRACE score or from richer trace features. As shown in Table~\ref{tab:trace_bridge}, calibrating the TRACE scalar improves Brier from $0.182$ to $0.154$ while leaving AUROC unchanged at $0.772$. Learning the three core operators further improves to $0.140$ Brier and $0.778$ AUROC, while the full TRACE+ reaches $0.137$ and $0.792$.
Thus, TRACE+ improves ranking as well as probability calibration, showing that its gains extend beyond scalar post-hoc calibration.

\begin{table}[t]
\centering
\footnotesize
\setlength{\tabcolsep}{2.2pt}
\renewcommand{\arraystretch}{1.05}
\begin{tabular}{@{}lcccc@{}}
\toprule
\multirow{2}{*}{Variant} &
\multicolumn{2}{c}{Main} &
\multicolumn{2}{c}{Cross} \\
\cmidrule(lr){2-3}
\cmidrule(lr){4-5}
& Brier $\downarrow$ & AUROC $\uparrow$
& Brier $\downarrow$ & AUROC $\uparrow$ \\
\midrule
TRACE+ & 0.137 & 0.792 & 0.120 & 0.817 \\
\quad + seq. likelihood & 0.139 & 0.787 & 0.121 & 0.816 \\
\quad w/o local entropy & 0.138 & 0.786 & 0.121 & 0.812 \\
\quad w/o local surprisal & 0.137 & 0.792 & 0.121 & 0.815 \\
\quad w/o trajectory slope & 0.136 & 0.792 & 0.123 & 0.808 \\
\quad w/o length & 0.151 & 0.757 & 0.127 & 0.792 \\
\midrule
Entropy-local only & 0.150 & 0.760 & 0.129 & 0.786 \\
Likelihood-local only & 0.144 & 0.770 & 0.129 & 0.782 \\
\bottomrule
\end{tabular}
\caption{
Ablation of TRACE+ components in main and cross-model settings,
reporting Brier and AUROC for feature removals and local-risk-only variants.
}
\label{tab:trace_ablation}
\end{table}

\paragraph{TRACE+ feature ablations.}
Table~\ref{tab:trace_ablation} evaluates TRACE+ feature variants in the main four-task and seven-model cross-model settings. Our final TRACE+ uses only trace-derived features, excluding conventional selected-token likelihood summaries. Adding minimum token probability, geometric mean token probability, and low-probability token rate slightly hurts performance, reducing main-setting AUROC from $0.792$ to $0.787$ and cross-model AUROC from $0.817$ to $0.816$. Feature-removal results further show that the trace signals are complementary. Entropy-only and likelihood-only variants are consistently weaker than TRACE+, while removing trajectory slope gives a marginal main-setting Brier gain but reduces cross-model AUROC from $0.817$ to $0.808$. We therefore retain the full trace-only feature set for stronger overall cross-model robustness.

\paragraph{Semantic fusion analysis.}
We further test whether localized trace risk is complementary to semantic reweighting.
As shown in Table~\ref{tab:semantic_fusion}, adding TRACE components improves MARS~\citep{bakman2024mars} from $0.173/0.754$ to $0.141/0.775$ Brier/AUROC and TokenSAR~\citep{duan2024shifting} from $0.174/0.738$ to $0.141/0.781$.
Conversely, adding MARS to TRACE+ slightly worsens both metrics, while TokenSAR raises AUROC only marginally from $0.7920$ to $0.7959$ without improving Brier.
Thus, localized trace risk strengthens semantic estimators, whereas semantic scores add little additional information beyond TRACE+.

\begin{table}[t]
\centering
\footnotesize
\setlength{\tabcolsep}{2.8pt}
\renewcommand{\arraystretch}{1.05}
\begin{tabular*}{\columnwidth}{@{\extracolsep{\fill}}lcc@{}}
\toprule
Method & Brier $\downarrow$ & AUROC $\uparrow$ \\
\midrule
MARS~\citep{bakman2024mars}
                  & 0.1730 & 0.7540 \\
MARS + TRACE      & 0.1410 & 0.7750 \\
TokenSAR~\citep{duan2024shifting}
                  & 0.1740 & 0.7380 \\
TokenSAR + TRACE  & 0.1410 & 0.7810 \\
\midrule
TRACE+            & 0.1371 & 0.7920 \\
TRACE+ + MARS     & 0.1376 & 0.7910 \\
TRACE+ + TokenSAR & 0.1374 & 0.7959 \\
\bottomrule
\end{tabular*}
\caption{
Semantic fusion between TRACE and the semantic confidence estimators MARS and TokenSAR.
}
\label{tab:semantic_fusion}
\end{table}

\begin{figure}[t]
    \centering
    \includegraphics[width=\columnwidth]{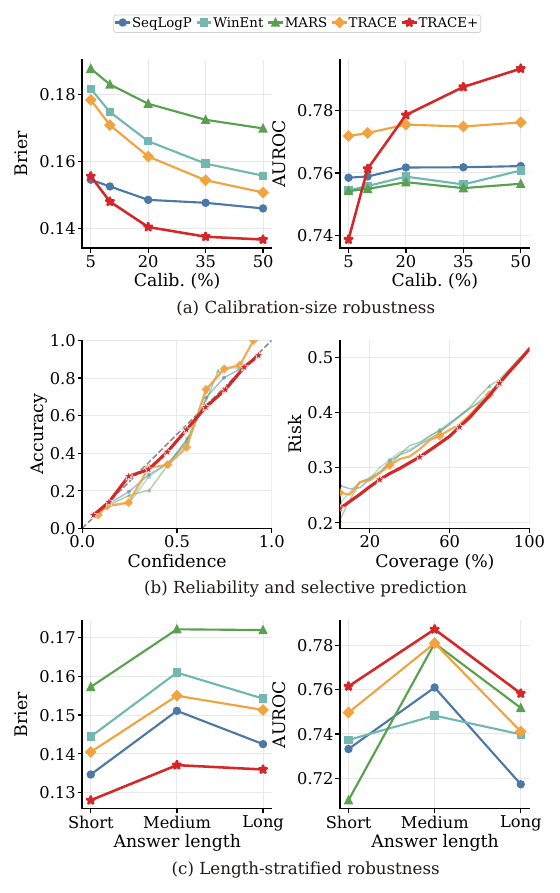}
    \caption{
    Robustness and reliability analyses.
    TRACE+ remains strong across calibration sizes, reliability/selective prediction, and answer-length strata.
    }
    \label{fig:analysis_six_panel}
\end{figure}

\paragraph{Calibration-size robustness.}
Figure~\ref{fig:analysis_six_panel} (a) evaluates how TRACE+ scales with the size of the held-out calibration split on MLQA, SVAMP, TriviaQA, and TruthfulQA. With only $5\%$ calibration data, TRACE+ already matches the strongest baseline in Brier score, although its AUROC is less stable. As the calibration split increases, TRACE+ improves consistently: from $20\%$ onward, it achieves both the lowest Brier score and the highest AUROC among all compared estimators. At the default $35\%$ split, TRACE+ obtains $0.137$ Brier and $0.792$ AUROC, outperforming SeqLogP / Total NLL at $0.148$ Brier and $0.762$ AUROC. This suggests that TRACE+ benefits from a modest calibration set, but does not require a large labeled split.

\paragraph{Reliability and selective prediction.}
Figure~\ref{fig:analysis_six_panel} (b) provides a complementary view of how the calibrated confidence scores behave. In the reliability diagram, TRACE+ stays close to the diagonal across confidence bins, indicating that its predicted confidence better matches empirical accuracy. The risk-coverage curve evaluates selective prediction by retaining the most confident answers at each coverage level under a rejection setting. TRACE+ yields lower risk over most coverage levels, showing that its confidence scores are useful not only as calibrated probabilities, but also for ranking which answers should be trusted in downstream use.

\paragraph{Length-stratified robustness.}
Figure~\ref{fig:analysis_six_panel} (c) tests whether TRACE+ mainly exploits answer-length effects. We split examples into short, medium, and long groups according to generated answer length and evaluate all methods under the same calibration protocol. TRACE+ maintains the lowest Brier score across all three groups and achieves the strongest AUROC in each group across length strata. The advantage remains clear for long answers, where sequence-likelihood scores are more sensitive to accumulated token probabilities. This indicates that TRACE+ is not simply learning a length proxy, but captures localized decoding-risk evidence that remains useful across lengths. Detailed results corresponding to Figure~\ref{fig:analysis_six_panel} are provided for completeness in Appendix~\ref{app:analysis_six_panel_data}.

\paragraph{Cross-task calibration transfer.}
We further evaluate a stricter transfer setting where the calibrator is trained on three source tasks and directly evaluated on a held-out target task without target-task calibration labels. As shown in Table~\ref{tab:cross_task_transfer_oracle}, TRACE+ preserves competitive correctness ranking: it slightly improves AUROC on MLQA ($0.814$ vs. $0.813$) and TruthfulQA ($0.659$ vs. $0.644$), while remaining close on SVAMP ($0.842$ vs. $0.843$) and TriviaQA ($0.832$ vs. $0.835$). However, Brier score degrades on tasks with substantially different correctness priors, especially MLQA ($0.347$ vs. $0.264$) and TruthfulQA ($0.212$ vs. $0.137$). This indicates that decoding-trace features transfer better as ranking signals than as fully calibrated probabilities, and supports our use of a small held-out target calibration split for TRACE+ in the main protocol.

\begin{table}[t]
\centering
\footnotesize
\setlength{\tabcolsep}{1.2pt}
\renewcommand{\arraystretch}{0.95}

\resizebox{\columnwidth}{!}{%
\begin{tabular}{@{}lcccccc@{}}
\toprule
\multirow{2}{*}{Target}
& \multicolumn{2}{c}{Best Non-TRACE}
& \multicolumn{2}{c}{TRACE+}
& \multicolumn{2}{c}{Gain} \\
\cmidrule(lr){2-3}
\cmidrule(lr){4-5}
\cmidrule(lr){6-7}
& Brier $\downarrow$
& AUROC $\uparrow$
& Brier $\downarrow$
& AUROC $\uparrow$
& Brier $\uparrow$
& AUROC $\uparrow$ \\
\midrule
MLQA       & 0.264 & 0.813 & 0.347 & 0.814 & -0.083 & +0.001 \\
SVAMP      & 0.144 & 0.843 & 0.140 & 0.842 & +0.004 & -0.001 \\
TriviaQA   & 0.182 & 0.835 & 0.169 & 0.832 & +0.013 & -0.003 \\
TruthfulQA & 0.137 & 0.644 & 0.212 & 0.659 & -0.075 & +0.016 \\
\bottomrule
\end{tabular}%
}
\caption{
Cross-task transfer against the strongest source-calibrated baseline on unseen held-out target tasks.
}
\label{tab:cross_task_transfer_oracle}
\end{table}

\section{Conclusion}

In this paper, we study general generation calibration and propose \textsc{TRACE}, a single-pass confidence estimator that models decoding-time uncertainty as a structured token-level trajectory. TRACE combines position-sensitive, local-peak, and length-normalized local risk operators to produce a label-free confidence score. TRACE+ learns richer trace features with a lightweight held-out calibrator to produce calibrated correctness probabilities. Across 19 baselines and eight LLMs, TRACE consistently improves correctness ranking over standard likelihood and entropy estimators, while TRACE+ further improves calibration and ranking, achieving the best overall Brier score and AUROC. Ablations show gains arise from localized decoding-trace evidence and remain robust across models, answer lengths, and calibration sizes.

\section*{Limitations}
TRACE requires token-level probabilities or entropy during decoding, which may be unavailable from closed-source APIs. Since TRACE estimates confidence from a single decoded answer, it complements retrieval, multi-sample consistency, and external verification for factual errors with little decoding-time instability. TRACE+ further requires a representative held-out calibration split.

\section*{Acknowledgments}
This work was supported by the National Natural Science Foundation of China (NSFC) under Grant No. 62306256 and the Natural Science Foundation of Guangdong Province under Grant No. 2025A1515010261.

\bibliography{custom}
\clearpage

\appendix
\label{sec:appendix}

\section*{Appendix Contents}

\noindent
\begin{tabular}{@{}ll@{}}
\hyperref[app:trace_spike_evidence]{{A}}
& \hyperref[app:trace_spike_evidence]{{Localized Risk Spike Analysis}} \\

\hyperref[app:trace_features]{{B}}
& \hyperref[app:trace_features]{{TRACE Feature Definitions}} \\

\hyperref[app:Protocol]{{C}}
& \hyperref[app:Protocol]{{Protocol Comparison}} \\

\hyperref[app:baseline_protocol_details]{{D}}
& \hyperref[app:baseline_protocol_details]{{Baseline and Protocol Details}} \\

\hyperref[app:supp_experiments]{{E}}
& \hyperref[app:supp_experiments]{{Supplementary Experimental Results}} \\

\quad \hyperref[app:broader_uq]{E.1}
& \hyperref[app:broader_uq]{Comparison with Broader UQ Methods} \\

\quad \hyperref[app:cross_model_details]{E.2}
& \hyperref[app:cross_model_details]{Detailed Cross-Model Results} \\

\quad \hyperref[app:localized_error_analysis]{E.3}
& \hyperref[app:localized_error_analysis]{Localized-Error Analysis} \\

\quad \hyperref[app:trace_operator_analysis]{E.4}
& \hyperref[app:trace_operator_analysis]{TRACE Operator Analysis} \\

\quad \hyperref[app:analysis_six_panel_data]{E.5}
& \hyperref[app:analysis_six_panel_data]{Robustness and Reliability Analyses} \\
\end{tabular}

\vspace{0.8em}

\section{Localized Risk Spike Analysis}
\label{app:trace_spike_evidence}

This section reports the tabular data and additional modeling results corresponding to Figure~\ref{fig:trace_spike_evidence} in the main paper. The goal is to make the localized-risk analysis fully reproducible and to quantify the trends visualized in the figure. For each decoded answer, we compute the mean token-level risk, which represents global aggregation, and a spike gap, defined as the excess of the largest local-risk region over the global mean. Table~\ref{tab:spike_strata_rates} reports the underlying strata used in the figure: examples are grouped by mean risk and then split into low- and high-spike groups. Within the same mean-risk stratum, high-spike examples consistently have higher incorrect rates, showing that localized spikes provide risk information beyond global mean aggregation.

\begin{table}[H]
\centering
\footnotesize
\setlength{\tabcolsep}{2.2pt}
\renewcommand{\arraystretch}{1.03}

\begin{tabular}{@{}llrrrr@{}}
\toprule
Mean risk & Spike gap & $n$ & Mean risk & Spike gap & Incorr. rate \\
\midrule
Low  & Low  & 313 & 0.064 & 0.023 & 0.224 \\
Low  & High & 312 & 0.090 & 0.182 & 0.554 \\
Mid  & Low  & 313 & 0.178 & 0.102 & 0.591 \\
Mid  & High & 313 & 0.189 & 0.390 & 0.843 \\
High & Low  & 313 & 0.377 & 0.139 & 0.850 \\
High & High & 312 & 0.321 & 0.454 & 0.904 \\
\bottomrule
\end{tabular}
\caption{Incorrect rates by mean risk and spike gap.}
\label{tab:spike_strata_rates}
\end{table}

Table~\ref{tab:spike_modeling_metrics} gives a complementary modeling view of the same phenomenon.
Mean risk alone provides a strong global summary, but adding spike-gap information improves AUROC from $0.774$ to $0.808$ and reduces Brier from $0.179$ to $0.164$.
The full TRACE local-feature set further improves to $0.818$ AUROC and $0.159$ Brier, confirming that the visual pattern in Figure~\ref{fig:trace_spike_evidence} corresponds to measurable predictive signal.

\begin{table}[H]
\centering
\footnotesize
\setlength{\tabcolsep}{3.2pt}
\renewcommand{\arraystretch}{1.03}

\begin{tabular}{@{}lcc@{}}
\toprule
Feature set & AUROC $\uparrow$ & Brier $\downarrow$ \\
\midrule
Mean risk & $0.774 \pm 0.015$ & $0.179 \pm 0.004$ \\
Max token risk & $0.784 \pm 0.014$ & $0.173 \pm 0.005$ \\
Spike gap & $0.761 \pm 0.018$ & $0.184 \pm 0.005$ \\
Mean risk + spike gap & $0.808 \pm 0.013$ & $0.164 \pm 0.005$ \\
TRACE local features
& $\mathbf{0.818 \pm 0.012}$
& $\mathbf{0.159 \pm 0.005}$ \\
\bottomrule
\end{tabular}

\caption{Predictive value of localized risk features.}
\label{tab:spike_modeling_metrics}
\end{table}

\section{TRACE Feature Definitions}
\label{app:trace_features}

This section summarizes the features used by TRACE and TRACE+.
For each decoded answer, we record the selected-token probability and the predictive entropy at each decoding step.
These values form two token-level traces: selected-token surprisal and token entropy.
TRACE estimates answer-level risk from the structure of this decoding trace, with emphasis on where uncertainty appears rather than only how large the average uncertainty is.

\paragraph{TRACE score.}
TRACE combines three fixed risk signals.
First, it uses a position-decayed entropy score, which gives more weight to uncertainty at earlier decoding steps.
Second, it uses the maximum entropy over a short local window, which preserves localized uncertainty spikes.
Third, it uses a length-normalized total surprisal term, which retains selected-token likelihood information without applying the full penalty of unnormalized sequence likelihood.
In all experiments, TRACE uses the same fixed configuration:

\begin{equation}
\begin{aligned}
R_{\mathrm{TRACE}} ={}& 0.40D_2(H)+0.40M_H^{(4)} \\
&+0.20L_{1/4}(s),
\end{aligned}
\end{equation}
where $D_{2}(H)$ is the position-decayed entropy score, $M_H^{(4)}$ is the maximum four-token entropy window, and $L_{1/4}(s)$ is the length-normalized surprisal term.
The raw TRACE confidence is $\exp(-R_{\textsc{TRACE}})$.
This score is used directly for ranking metrics; for Brier score and ECE, it is calibrated with the same held-out protocol as the baselines.

\paragraph{TRACE+ features.}
TRACE+ uses a lightweight logistic calibrator over trace-derived features.
Table~\ref{tab:trace_plus_features} summarizes the feature groups.
The final TRACE+ variant removes conventional selected-token likelihood summaries, so the reported TRACE+ results use only trace-localization features.
The last three rows in Table~\ref{tab:trace_plus_features} are included only in the ``+ seq. likelihood'' ablation.

\begin{table}[H]
\centering
\footnotesize
\setlength{\tabcolsep}{3.0pt}
\renewcommand{\arraystretch}{1.03}

\begin{tabular}{@{}lp{0.62\columnwidth}@{}}
\toprule
Group & Features \\
\midrule
Early entropy
    & First-3 entropy confidence \\
Global entropy
    & Mean entropy confidence \\
Length
    & Log answer length \\
Trajectory
    & Entropy slope \\
Local entropy
    & Decayed entropy confidence ($\lambda{=}2,4$) \\
Local surprisal
    & Decayed surprisal confidence ($\lambda{=}2,4$) \\
\midrule
Excluded likelihood
    & Minimum token probability \\
Excluded likelihood
    & Geometric mean token probability \\
Excluded likelihood
    & Low-probability token rate \\
\bottomrule
\end{tabular}

\caption{TRACE+ feature groups and definitions.}
\label{tab:trace_plus_features}
\end{table}

\section{Protocol comparison}
\label{app:Protocol}

This section summarizes the inference assumptions of representative confidence estimators.
Table~\ref{tab:protocol_comparison} compares whether each method scores the original decoded answer, can be computed from a single generation pass, requires additional generations, uses a semantic modeling component, or learns a held-out calibrator.
Likelihood- and entropy-based scores are inexpensive and single-pass, but they typically collapse the decoding trace into global statistics.
Beam-based scores still preserve the evaluated answer, but require additional beam statistics from the same prompt.
Sampling-based methods such as SelfCheckGPT and Semantic Entropy can capture uncertainty beyond a single decoded answer, but require extra generations and may no longer score exactly the same answer under the same inference workflow.
TRACE is designed for the stricter single-pass, decoded-answer-preserving setting, while TRACE+ adds only a lightweight held-out calibrator for calibrated probabilities.
\begin{table}[t]
\centering
\footnotesize
\setlength{\tabcolsep}{3.0pt}
\renewcommand{\arraystretch}{1.10}

\begin{tabularx}{\columnwidth}{@{}Xccccc@{}}
\toprule
Method & Same & Single & Extra & Sem. & Calib. \\
\midrule
Likelihood / entropy & \cmark & \cmark & \xmark & \xmark & \xmark \\
Beam-based scores & \cmark & \pmark & \xmark & \xmark & \xmark \\
TokenSAR & \cmark & \cmark & \xmark & \cmark & \xmark \\
MARS & \cmark & \cmark & \xmark & \cmark & \xmark \\
SelfCheckGPT & \cmark & \xmark & \cmark & \pmark & \xmark \\
Semantic Entropy & \xmark & \xmark & \cmark & \cmark & \xmark \\
\midrule
TRACE & \cmark & \cmark & \xmark & \xmark & \xmark \\
TRACE+ & \cmark & \cmark & \xmark & \xmark & \cmark \\
\bottomrule
\end{tabularx}
\caption{
Protocol comparison of confidence estimators by answer preservation,
single-pass inference, extra generations, semantic modeling, and calibration.
}
\label{tab:protocol_comparison}
\end{table}

\section{Baseline and Protocol Details}
\label{app:baseline_protocol_details}

\begingroup
\small

We compare TRACE+ against a broad pool of single-pass and near-single-pass confidence estimators.
For a generated answer of length $T$, let $p_t$ denote the probability assigned to the selected token at decoding step $t$, and let $H_t$ denote the predictive entropy at that step.
All likelihood- and entropy-based baselines are computed from the same generation trace used by TRACE.

\paragraph{Token- and sequence-likelihood baselines.}
\begin{itemize}[leftmargin=1.15em,itemsep=2pt,topsep=2pt,parsep=0pt]

    \item \textbf{First-Token Prob.} Probability assigned to the first generated token~\citep{chen2026firsttoken}.

    \item \textbf{MeanProb.} Arithmetic mean of selected-token probabilities over the generated answer~\citep{yona2022useful,flores2025improving}.

    \item \textbf{Len-Norm LogP.} Length-normalized sequence likelihood,
    $\exp\!\left(\frac{1}{T}\sum_{t=1}^{T}\log p_t\right)$~\citep{yona2022useful,bakman2024mars}.

    \item \textbf{SeqLogP / Total NLL.} Unnormalized sequence likelihood,
    $\exp\!\left(\sum_{t=1}^{T}\log p_t\right)$, which preserves answer-length effects~\citep{aichberger2024rethinking}.

    \item \textbf{MinProb.} Minimum selected-token probability over the generated answer~\citep{manakul2023selfcheckgpt}.

    \item \textbf{High-Prob Token Rate.} Fraction of generated tokens whose selected-token probability exceeds
a fixed high-confidence threshold, adapted from probability-bin
trajectory summaries~\citep{szymczyk2026monitoring}.
\end{itemize}

\paragraph{Entropy- and position-based baselines.}
\begin{itemize}[leftmargin=1.15em,itemsep=2pt,topsep=2pt,parsep=0pt]
    \item \textbf{Mean Token Entropy.} Negative mean token entropy over the generated answer~\citep{yona2022useful,flores2025improving}.

    \item \textbf{Max Token Entropy.} Negative maximum token entropy over the generated answer~\citep{manakul2023selfcheckgpt}.

    \item \textbf{First-$k$ Entropy.} Negative mean entropy over the first $k$ generated positions,
extending last-token entropy used in prior uncertainty estimation~\citep{chen2026firsttoken}.

    \item \textbf{Last-$k$ Entropy.} Negative mean entropy over the last $k$ generated positions~\citep{li2026semantic}.
\end{itemize}

\paragraph{Trajectory-diagnostic baselines.}
\begin{itemize}[leftmargin=1.15em,itemsep=2pt,topsep=2pt,parsep=0pt]
    \item \textbf{Windowed Entropy.} Minimum local-window uncertainty score along the decoding trajectory~\citep{sriramanan2024llm}.

    \item \textbf{Entropy Variance.} Negative variance of token entropies along the decoding trajectory~\citep{zhu2026edis}.

    \item \textbf{Entropy Slope.} Negative linear slope of token entropy over decoding positions~\citep{zhu2026edis}.

    \item \textbf{Token-Prob Slope.} Linear slope of selected-token probabilities over decoding positions,
motivated by temporal log-probability modeling~\citep{shapiro2026halt}.
\end{itemize}

\paragraph{Beam-distribution baselines.}
\begin{itemize}[leftmargin=1.15em,itemsep=2pt,topsep=2pt,parsep=0pt]
    \item \textbf{Beam-Ratio.} Probability gap between the best beam and lower-ranked beams~\citep{flores2025improving}.

    \item \textbf{Beam-TailThinness.} Concentration of normalized beam-level sequence probabilities~\citep{flores2025improving}.

    \item \textbf{Beam Entropy.} Negative entropy of the normalized beam-level sequence distribution~\citep{flores2025improving}.
\end{itemize}

\paragraph{Semantic-reweighting baselines.}
\begin{itemize}[leftmargin=1.15em,itemsep=2pt,topsep=2pt,parsep=0pt]
    \item \textbf{TokenSAR.} Token-likelihood aggregation reweighted by token semantic relevance~\citep{duan2024shifting}.

    \item \textbf{MARS.} Meaning-aware token-likelihood aggregation using semantic contribution weights~\citep{bakman2024mars}.
\end{itemize}

\paragraph{Scoring and calibration protocol.}
For AUROC, we use the raw score direction after converting uncertainty values into confidence scores when necessary.
For Brier score and ECE, each scalar baseline score is mapped to a calibrated probability using the same held-out calibration split as TRACE+.
Beam-Ratio, Beam-TailThinness, and Beam Entropy rely on beam-level output-distribution statistics from the same prompt, and are therefore \emph{near-single-pass} rather than strictly single-pass.
TokenSAR and MARS use semantic relevance or contribution weights to reweight token-level likelihood signals.

\paragraph{Protocol assumptions.}
Table~\ref{tab:protocol_comparison} compares representative confidence-estimation protocols along five inference assumptions:
\begin{itemize}[leftmargin=1.15em,itemsep=2pt,topsep=2pt,parsep=0pt]
    \item \textbf{Same} indicates whether a method scores the original decoded answer without replacing it.
    \item \textbf{Single} indicates whether the method can be computed from a single decoding pass.
    \item \textbf{Extra} indicates whether additional generations are required.
    \item \textbf{Sem.} indicates whether the method uses a semantic module, such as semantic relevance scoring, semantic equivalence grouping, or semantic consistency checking.
    \item \textbf{Calib.} indicates whether the method itself learns a held-out calibrator.
\end{itemize}

\paragraph{Protocol comparison details.}
\begin{itemize}[leftmargin=1.15em,itemsep=2pt,topsep=2pt,parsep=0pt]
    \item \textbf{Likelihood and entropy baselines.}
    These methods are single-pass and answer-preserving, but reduce the decoding process to global token- or sequence-level statistics~\citep{yona2022useful,aichberger2024rethinking}.

    \item \textbf{Beam-distribution methods.}
    These methods also preserve the evaluated answer, but require beam-level statistics from the same prompt; we therefore mark them as \emph{near-single-pass}~\citep{flores2025improving}.

    \item \textbf{Semantic reweighting methods.}
    TokenSAR~\citep{duan2024shifting} and MARS~\citep{bakman2024mars} remain answer-preserving, but use semantic relevance or meaning-contribution estimates to reweight token-level likelihood signals.

    \item \textbf{Sampling-based methods.}
    SelfCheckGPT requires additional generations and may optionally use semantic comparison across sampled responses~\citep{manakul2023selfcheckgpt}.

    \item \textbf{Semantic Entropy.}
    Semantic Entropy requires multiple generations and semantic equivalence modeling, and therefore does not operate on a single fixed decoded answer~\citep{kuhn2023semantic}.
\end{itemize}

\begin{table*}[t]
\centering
\scriptsize
\setlength{\tabcolsep}{2.3pt}
\renewcommand{\arraystretch}{1.04}

\resizebox{\textwidth}{!}{%
\begin{tabular}{@{}ll lccc lccc@{}}
\toprule
\multirow{2}{*}{Model} &
\multirow{2}{*}{Task} &
\multicolumn{4}{c}{Brier comparison} &
\multicolumn{4}{c}{AUROC comparison} \\
\cmidrule(lr){3-6}
\cmidrule(l){7-10}
& & Best estimator & Base $\downarrow$ & TRACE+ $\downarrow$ & $\Delta$ $\uparrow$
& Best estimator & Base $\uparrow$ & TRACE+ $\uparrow$ & $\Delta$ $\uparrow$ \\
\midrule
Llama-3.1-8B & SVAMP   & High-Prob Token Rate & 0.134 & 0.064 & +0.071 & Mean Token Entropy & 0.939 & 0.974 & +0.035 \\
             & TriviaQA& First-Token Prob.    & 0.161 & 0.158 & +0.003 & First-Token Prob.  & 0.798 & 0.801 & +0.003 \\
             & TQA-Gen & First-Token Prob.    & 0.088 & 0.085 & +0.003 & First-Token Prob.  & 0.646 & 0.683 & +0.037 \\
\addlinespace[1pt]
Mistral-7B   & SVAMP   & SeqLogP / Total NLL  & 0.190 & 0.178 & +0.012 & TokenSAR           & 0.824 & 0.808 & -0.016 \\
             & TriviaQA& First-Token Prob.    & 0.180 & 0.165 & +0.015 & Max Token Entropy  & 0.799 & 0.823 & +0.024 \\
             & TQA-Gen & Windowed Entropy     & 0.104 & 0.096 & +0.008 & First-$k$ Entropy  & 0.694 & 0.739 & +0.046 \\
\addlinespace[1pt]
Phi-3.5-MoE  & SVAMP   & SeqLogP / Total NLL  & 0.119 & 0.105 & +0.014 & SeqLogP / Total NLL& 0.923 & 0.928 & +0.004 \\
             & TriviaQA& First-Token Prob.    & 0.190 & 0.174 & +0.016 & First-Token Prob.  & 0.761 & 0.784 & +0.024 \\
             & TQA-Gen & First-$k$ Entropy    & 0.089 & 0.084 & +0.005 & First-$k$ Entropy  & 0.743 & 0.768 & +0.025 \\
\addlinespace[1pt]
Qwen2-57B    & SVAMP   & Beam-TailThinness    & 0.174 & 0.154 & +0.020 & Max Token Entropy  & 0.567 & 0.848 & +0.281 \\
             & TriviaQA& MinProb              & 0.185 & 0.180 & +0.006 & Max Token Entropy  & 0.781 & 0.780 & -0.001 \\
             & TQA-Gen & SeqLogP / Total NLL  & 0.099 & 0.096 & +0.004 & SeqLogP / Total NLL& 0.821 & 0.828 & +0.007 \\
\addlinespace[1pt]
Llama-3.1-70B& SVAMP   & SeqLogP / Total NLL  & 0.095 & 0.085 & +0.010 & Last-$k$ Entropy   & 0.895 & 0.889 & -0.007 \\
             & TriviaQA& SeqLogP / Total NLL  & 0.116 & 0.108 & +0.008 & SeqLogP / Total NLL& 0.842 & 0.853 & +0.011 \\
             & TQA-Gen & SeqLogP / Total NLL  & 0.102 & 0.100 & +0.003 & SeqLogP / Total NLL& 0.773 & 0.789 & +0.016 \\
\addlinespace[1pt]
Qwen3-32B    & SVAMP   & MARS                 & 0.020 & 0.017 & +0.003 & Entropy Slope      & 0.836 & 0.871 & +0.035 \\
             & TriviaQA& First-Token Prob.    & 0.212 & 0.207 & +0.005 & First-Token Prob.  & 0.720 & 0.728 & +0.007 \\
             & TQA-Gen & First-Token Prob.    & 0.107 & 0.099 & +0.007 & First-Token Prob.  & 0.736 & 0.773 & +0.037 \\
\addlinespace[1pt]
Gemma-2-9B   & SVAMP   & Beam-TailThinness    & 0.161 & 0.155 & +0.006 & SeqLogP / Total NLL& 0.828 & 0.851 & +0.023 \\
             & TriviaQA& SeqLogP / Total NLL  & 0.140 & 0.132 & +0.007 & SeqLogP / Total NLL& 0.878 & 0.878 & -0.001 \\
             & TQA-Gen & SeqLogP / Total NLL  & 0.090 & 0.089 & +0.000 & SeqLogP / Total NLL& 0.748 & 0.768 & +0.020 \\
\bottomrule
\end{tabular}%
}
\caption{Task-level details for cross-model generalization.}
\label{tab:cross_model_task_details}
\end{table*}

\paragraph{TRACE protocol.}
\begin{itemize}[leftmargin=1.15em,itemsep=2pt,topsep=2pt,parsep=0pt]
    \item \textbf{TRACE.}
    TRACE differs from the above alternatives by using only the token-level decoding trace of the original answer: selected-token surprisal, predictive entropy, and localized trajectory features.
    It does not require additional samples, semantic modules, external verifiers, or LLM-as-a-judge labels.

    \item \textbf{TRACE+.}
    TRACE+ uses the same trace-only features at inference time, but learns a lightweight held-out calibrator to map trace-derived risk patterns into calibrated answer-level probabilities.

    \item \textbf{Calibration fairness.}
    For fair comparison on probability metrics, scalar baselines are also mapped through the same held-out calibration protocol when computing Brier score and ECE.
\end{itemize}

\endgroup

\section{Supplementary Experimental Results}
\label{app:supp_experiments}

\subsection{Comparison with Broader UQ Methods}
\label{app:broader_uq}

We additionally compare TRACE and TRACE+ with uncertainty-estimation methods beyond token-likelihood, entropy, and semantic-reweighting baselines. These include LARS~\citep{yaldiz2025}, P(True)~\citep{kadavath2022language}, verbalized confidence~\citep{tian2023just}, and an internal-state probe~\citep{ji2024}. 
\begin{table}[t]
\centering
\small
\renewcommand{\arraystretch}{1.03}
\begin{tabular*}{\columnwidth}{@{\extracolsep{\fill}}lccc@{}}
\toprule
Method & Req. & Brier $\downarrow$ & AUROC $\uparrow$ \\
\midrule
MARS                 & --  & 0.173 & 0.754 \\
TokenSAR             & --  & 0.174 & 0.738 \\
LARS                 & L   & 0.157 & 0.758 \\
P(True)              & E   & 0.178 & 0.715 \\
Verbalized conf.     & E   & 0.183 & 0.697 \\
Internal-state probe & H+L & 0.159 & 0.766 \\
TRACE                & --  & 0.154 & 0.772 \\
TRACE+               & C   & \textbf{0.137} & \textbf{0.792} \\
\bottomrule
\end{tabular*}
\caption{Comparison with uncertainty-estimation methods for answer-level confidence and ranking.}
\label{tab:broader_uq}
\end{table}

Table~\ref{tab:broader_uq} reports average performance over MLQA, SVAMP, TriviaQA, and TruthfulQA, together with the additional requirements of each method: E denotes extra inference, H hidden-state access, L labeled training, and C held-out calibration labels. TRACE achieves the best AUROC without extra inference or training, while TRACE+ achieves the best overall Brier and AUROC with held-out calibration labels.

\subsection{Detailed Cross-Model Results}
\label{app:cross_model_details}

Table~\ref{tab:cross_model_best_estimators} reports the non-TRACE estimators that define the Best Non-TRACE columns in Table~\ref{tab:model_generalization}.
The best Brier and best AUROC estimators are selected independently for each target model after averaging over SVAMP, TriviaQA, and TruthfulQA.
Table~\ref{tab:cross_model_task_details} further expands the task-level comparisons used to compute the Task Wins columns.
\begin{table*}[t]
\centering
\scriptsize
\setlength{\tabcolsep}{3.2pt}
\renewcommand{\arraystretch}{1.06}

\resizebox{\textwidth}{!}{%
\begin{tabular}{@{}llcc lcc cc@{}}
\toprule
\multirow{2}{*}{Model} &
\multicolumn{3}{c}{Best Non-TRACE for Brier} &
\multicolumn{3}{c}{Best Non-TRACE for AUROC} &
\multicolumn{2}{c}{TRACE+} \\
\cmidrule(lr){2-4}
\cmidrule(lr){5-7}
\cmidrule(l){8-9}
& Estimator & Brier $\downarrow$ & $\Delta$Brier $\uparrow$
& Estimator & AUROC $\uparrow$ & $\Delta$AUROC $\uparrow$
& Brier $\downarrow$ & AUROC $\uparrow$ \\
\midrule
Llama-3.1-8B-Instruct      & High-Prob Token Rate & 0.141 & +0.039 & Mean Token Entropy & 0.747 & +0.072 & 0.102 & 0.819 \\
Mistral-7B-Instruct-v0.3   & SeqLogP / Total NLL  & 0.160 & +0.014 & TokenSAR           & 0.763 & +0.027 & 0.146 & 0.790 \\
Phi-3.5-MoE-Instruct       & SeqLogP / Total NLL  & 0.135 & +0.014 & SeqLogP / Total NLL& 0.793 & +0.034 & 0.121 & 0.827 \\
Qwen2-57B-A14B             & SeqLogP / Total NLL  & 0.169 & +0.026 & Max Token Entropy  & 0.714 & +0.105 & 0.143 & 0.819 \\
Llama-3.1-70B-Instruct     & SeqLogP / Total NLL  & 0.104 & +0.007 & SeqLogP / Total NLL& 0.819 & +0.024 & 0.098 & 0.844 \\
Qwen3-32B                  & First-Token Prob.    & 0.113 & +0.005 & SeqLogP / Total NLL& 0.693 & +0.097 & 0.108 & 0.791 \\
Gemma-2-9B                 & SeqLogP / Total NLL  & 0.133 & +0.007 & SeqLogP / Total NLL& 0.818 & +0.014 & 0.126 & 0.832 \\
\bottomrule
\end{tabular}%
}
\caption{Best averaged non-TRACE estimators in the cross-model study.}
\label{tab:cross_model_best_estimators}
\end{table*}

\begin{table*}[t]
\centering
{
\small
\setlength{\tabcolsep}{3.0pt}

\begin{tabular*}{\textwidth}{@{\extracolsep{\fill}}lccccccc@{}}
\toprule
Method
& \multicolumn{2}{c}{AUROC $\uparrow$}
& \multicolumn{4}{c}{Localized errors}
& \multicolumn{1}{c}{Global errors} \\
\cmidrule(lr){2-3}
\cmidrule(lr){4-7}
\cmidrule(l){8-8}
& Loc.
& Glob.
& TPR@5\% $\uparrow$
& TPR@10\% $\uparrow$
& TPR@20\% $\uparrow$
& pAUC@10\% $\uparrow$
& TPR@10\% $\uparrow$ \\
\midrule
SeqLogP
& 0.763 & 0.767 & 0.190 & 0.395 & 0.555 & 0.572 & 0.330 \\
WindowEnt
& 0.778 & 0.754 & 0.245 & 0.378 & 0.664 & 0.599 & 0.371 \\
MARS
& 0.778 & 0.789 & 0.308 & 0.494 & 0.637 & 0.620 & 0.330 \\
TRACE
& 0.796 & \textbf{0.796} & \textbf{0.346} & 0.471 & 0.691 & 0.618 & \textbf{0.474} \\
TRACE+
& \textbf{0.805} & 0.744 & 0.272 & \textbf{0.521} & \textbf{0.692} & \textbf{0.648} & 0.423 \\
\bottomrule
\end{tabular*}
\caption{
Localized- and global-error discrimination.
}
\label{tab:localized_global_errors}

}
\end{table*}

\subsection{Localized-Error Analysis}
\label{app:localized_error_analysis}

We further analyze whether decoding-trace localization is particularly useful for errors whose correctness depends on a compact semantic span. We divide incorrect generations into two groups. \emph{Localized errors} are failures determined by a compact part of the answer, including number/arithmetic errors, incorrect entities or spans, and unsupported factual claims. \emph{Global errors} include off-topic, incoherent, abstaining, or multi-span failures. For each error slice, we compare against correct answers matched by task and answer length.

\paragraph{Localized versus global errors.}
Table~\ref{tab:localized_global_errors} reports discrimination performance for localized and global errors. TRACE and TRACE+ improve localized-error AUROC from $0.778$ for the strongest non-TRACE baseline to $0.796$ and $0.805$, respectively. The advantage also holds at low-FPR operating points: TRACE performs best at $5\%$ FPR, while TRACE+ achieves the highest TPR at $10\%$ and $20\%$ FPR and the strongest partial AUROC below $10\%$ FPR. For global errors, the fixed TRACE score achieves the highest AUROC among the compared methods. These results suggest that preserving localized decoding uncertainty is particularly useful for compact-span failures, while the fixed TRACE aggregation also remains effective for more diffuse generation errors.

\paragraph{Localized-error subtypes.}
We decompose the localized set into number/arithmetic, entity/span, and factual-claim errors. As shown in Table~\ref{tab:localized_error_subtypes}, TRACE+ achieves the strongest AUROC on all three subtypes, reaching $0.927$, $0.836$, and $0.607$, respectively. Compared with the strongest non-TRACE baseline for each subtype, these correspond to gains of $0.048$, $0.035$, and $0.018$. The improvement is largest for number/arithmetic and entity/span errors, where correctness often turns on a short critical token span. The trend persists for factual-claim errors despite a smaller subset.

\subsection{TRACE Operator Analysis}
\label{app:trace_operator_analysis}

This section provides additional analysis of the three operators used in the fixed TRACE score: position-decayed entropy $D$, local-window entropy $M$, and length-normalized surprisal $L$. We examine their individual contributions, the importance of preserving positional structure, and robustness to the fixed hyperparameter configuration.

\paragraph{Operator contributions.}
Table~\ref{tab:trace_operator_ablation} compares each operator individually with their fixed and learned combinations. The three individual operators obtain similar but complementary discrimination performance, with average AUROC ranging from $0.751$ to $0.758$. Combining them in TRACE improves AUROC to $0.772$, while a learned combination of the same operator outputs reaches $0.776$. Thus, the fixed TRACE aggregation captures most of the benefit of combining the three risk signals without task-specific fitting.

\begin{table}[t]
\centering
\footnotesize
\renewcommand{\arraystretch}{1.03}

\begin{tabular*}{\columnwidth}{@{\extracolsep{\fill}}lc@{}}
\toprule
Variant & Avg. AUROC $\uparrow$ \\
\midrule
Position-decayed entropy $D$    & 0.751 \\
Local-window entropy $M$        & 0.754 \\
Length-normalized surprisal $L$ & 0.758 \\
\midrule
TRACE $(D+M+L)$                 & \underline{0.772} \\
Learned $(D,M,L)$               & \textbf{0.776} \\
\bottomrule
\end{tabular*}

\caption{Operator-level analysis of TRACE components and their fusion.}
\label{tab:trace_operator_ablation}
\end{table}

\paragraph{Position sensitivity.}
We further test whether TRACE benefits from where risk occurs along the decoding trajectory rather than only from the overall distribution of token-level risk. We perturb the positions of the risk signals while preserving their values and recompute the TRACE ranking score. As shown in Table~\ref{tab:trace_position_shift}, this perturbation consistently reduces AUROC across all four tasks, lowering the average from $0.772$ to $0.755$. The result indicates that the ordering and location of uncertainty provide useful information beyond the marginal magnitude of token-level risk.

\paragraph{Hyperparameter sensitivity.}
We evaluate the robustness of TRACE to its fixed hyperparameters by varying one parameter at a time. When varying one mixture weight, the other two weights are renormalized according to their original ratio. As shown in Table~\ref{tab:trace_hyperparameter_sensitivity}, performance remains stable across broad parameter ranges. Across all tested settings, Brier varies from $0.152$ to $0.160$ and AUROC from $0.763$ to $0.775$, indicating that TRACE does not rely on a narrow hyperparameter configuration.

\begin{table}[t]
\centering
\footnotesize
\setlength{\tabcolsep}{2.5pt}

\begin{tabular}{@{}lccc@{}}
\toprule
Param. & Values tested & Brier $\downarrow$ & AUROC $\uparrow$ \\
\midrule
$\alpha$  & $0.2$--$0.6$             & $0.154$--$0.155$ & $0.767$--$0.775$ \\
$\beta$   & $0.2$--$0.6$             & $0.154$--$0.156$ & $0.767$--$0.774$ \\
$\gamma$  & $0.1$--$0.4$             & $0.153$--$0.156$ & $0.769$--$0.772$ \\
$\lambda$ & $1,2,4,8$                & $0.154$--$0.156$ & $0.763$--$0.774$ \\
$w$       & $2,4,6,8$                & $0.153$--$0.156$ & $0.769$--$0.774$ \\
$\rho$    & $0,.125,.25,.5,.75,1$    & $0.152$--$0.160$ & $0.766$--$0.773$ \\
\bottomrule
\end{tabular}
\caption{Hyperparameter sensitivity analysis.}
\label{tab:trace_hyperparameter_sensitivity}
\end{table}

\begin{table}[t]
\centering
\footnotesize
\renewcommand{\arraystretch}{1.03}

\begin{tabular*}{\columnwidth}{@{\extracolsep{\fill}}lcc@{}}
\toprule
Task & Original $\uparrow$ & Position-shifted $\uparrow$ \\
\midrule
MLQA       & 0.786 & 0.769 \\
SVAMP      & 0.825 & 0.809 \\
TriviaQA   & 0.835 & 0.811 \\
TruthfulQA & 0.641 & 0.630 \\
\midrule
Average    & \textbf{0.772} & 0.755 \\
\bottomrule
\end{tabular*}

\caption{AUROC under risk-position perturbation.}
\label{tab:trace_position_shift}
\end{table}

\begin{table}[t]
\centering
\small
\setlength{\tabcolsep}{3.0pt}
\begin{tabular}{lcccc}
\toprule
Method
& \shortstack{Number/\\Arithmetic}
& \shortstack{Entity/\\Span}
& \shortstack{Factual\\Claim}
& \shortstack{Overall\\Localized} \\
& $(n=92)$
& $(n=438)$
& $(n=65)$
& $(n=595)$ \\
\midrule
SeqLogP
& 0.845 & 0.801 & 0.528 & 0.757 \\
WindowEnt
& 0.868 & 0.801 & 0.556 & 0.770 \\
MARS
& 0.879 & 0.796 & 0.589 & 0.768 \\
TRACE
& 0.871 & 0.821 & 0.569 & 0.784 \\
TRACE+
& \textbf{0.927}
& \textbf{0.836}
& \textbf{0.607}
& \textbf{0.805} \\
\bottomrule
\end{tabular}
\caption{AUROC on localized-error subtypes.}
\label{tab:localized_error_subtypes}
\end{table}

\subsection{Robustness and Reliability Analyses}
\label{app:analysis_six_panel_data}

This section provides the tabular data underlying Figure~\ref{fig:analysis_six_panel}. We test whether TRACE+ remains stable under three conditions: calibration-set size, calibrated confidence and selective prediction, and answer length. Results are averaged over MLQA, SVAMP, TriviaQA, and TruthfulQA using the main evaluation split protocol. Table~\ref{tab:calibration_size_data} reports performance as the calibration split increases from $5\%$ to $50\%$, testing whether TRACE+ requires a labeled calibration set. Table~\ref{tab:reliability_selective_data} reports ECE, Brier, AUROC, and selective-prediction risk at coverage levels, evaluating confidence reliability and its usefulness for deciding which answers to trust. For Table~\ref{tab:reliability_selective_data}, metrics here are computed after pooling predictions across tasks within each split. Table~\ref{tab:length_stratified_data} reports performance on short, medium, and long answers, testing whether TRACE+ remains effective across generation lengths.

\begin{table}[]
\centering
\footnotesize
\setlength{\tabcolsep}{1.4pt}
\renewcommand{\arraystretch}{1.02}

\resizebox{\columnwidth}{!}{%
\begin{tabular}{@{}llccccc@{}}
\toprule
Calib. & Metric & SeqLogP & WinEnt & MARS & TRACE & TRACE+ \\
\midrule

\multirow{2}{*}{5\%}
& Brier & 0.154 & 0.182 & 0.188 & 0.178 & 0.156 \\
& AUROC & 0.758 & 0.754 & 0.754 & 0.772 & 0.739 \\
\cmidrule(lr){1-7}

\multirow{2}{*}{10\%}
& Brier & 0.152 & 0.175 & 0.183 & 0.171 & 0.148 \\
& AUROC & 0.759 & 0.756 & 0.755 & 0.773 & 0.761 \\
\cmidrule(lr){1-7}

\multirow{2}{*}{20\%}
& Brier & 0.148 & 0.166 & 0.177 & 0.161 & 0.140 \\
& AUROC & 0.762 & 0.759 & 0.757 & 0.775 & 0.778 \\
\cmidrule(lr){1-7}

\multirow{2}{*}{35\%}
& Brier & 0.148 & 0.159 & 0.172 & 0.154 & 0.137 \\
& AUROC & 0.762 & 0.756 & 0.755 & 0.775 & 0.792 \\
\cmidrule(lr){1-7}

\multirow{2}{*}{50\%}
& Brier & 0.146 & 0.156 & 0.170 & 0.151 & 0.137 \\
& AUROC & 0.762 & 0.761 & 0.757 & 0.776 & 0.793 \\
\bottomrule
\end{tabular}%
}
\caption{Calibration-size robustness in Brier/AUROC.}
\label{tab:calibration_size_data}
\end{table}

\begin{table}[]
\centering
\footnotesize
\setlength{\tabcolsep}{1.3pt}
\renewcommand{\arraystretch}{1.02}

\resizebox{\columnwidth}{!}{%
\begin{tabular}{@{}lcccccc@{}}
\toprule
Method & ECE $\downarrow$ & Brier $\downarrow$ & AUROC $\uparrow$
& R@10 $\downarrow$ & R@50 $\downarrow$ & R@90 $\downarrow$ \\
\midrule
SeqLogP & 0.036 & 0.146 & 0.826 & 0.262 & 0.352 & 0.474 \\
WinEnt  & 0.058 & 0.149 & 0.826 & 0.256 & 0.357 & 0.476 \\
MARS    & 0.048 & 0.159 & 0.783 & 0.242 & 0.357 & 0.479 \\
TRACE   & 0.059 & 0.144 & 0.830 & 0.251 & 0.350 & 0.473 \\
TRACE+  & 0.013 & 0.134 & 0.877 & 0.239 & 0.331 & 0.474 \\
\bottomrule
\end{tabular}%
}
\caption{Reliability and selective-prediction results.}
\label{tab:reliability_selective_data}
\end{table}

\begin{table}[H]
\centering
\footnotesize
\renewcommand{\arraystretch}{1.03}

\begin{tabular*}{\columnwidth}{@{\extracolsep{\fill}}lccc@{}}
\toprule
Method & Short & Medium & Long \\
\midrule
SeqLogP & 0.135/0.733 & 0.151/0.761 & 0.142/0.717 \\
WinEnt  & 0.144/0.737 & 0.161/0.748 & 0.154/0.740 \\
MARS    & 0.157/0.710 & 0.172/0.781 & 0.172/0.752 \\
TRACE   & 0.140/0.750 & 0.155/0.781 & 0.151/0.741 \\
TRACE+  & 0.128/0.761 & 0.137/0.792 & 0.136/0.758 \\
\bottomrule
\end{tabular*}

\caption{Length-stratified robustness in Brier/AUROC.}
\label{tab:length_stratified_data}
\end{table}

These results show that TRACE+ is robust across evaluated calibration sizes, produces reliable confidence scores after calibration, and remains effective across answer-length strata. They also provide the values corresponding to the curves in Figure~\ref{fig:analysis_six_panel}, facilitating robustness and reliability analyses.

\end{document}